# Learning Inclusion-Optimal Chordal Graphs

**Vincent Auvray**
GIGA-R and EE & CS Dept.
University of Liège
Vincent.Auvray@ulg.ac.be

**Louis Wehenkel**
GIGA-R and EE & CS Dept.
University of Liège
L.Wehenkel@ulg.ac.be

## Abstract

Chordal graphs can be used to encode dependency models that are representable by both directed acyclic and undirected graphs. This paper discusses a very simple and efficient algorithm to learn the chordal structure of a probabilistic model from data. The algorithm is a greedy hill-climbing search algorithm that uses the inclusion boundary neighborhood over chordal graphs. In the limit of a large sample size and under appropriate hypotheses on the scoring criterion, we prove that the algorithm will find a structure that is inclusion-optimal when the dependency model of the data-generating distribution can be represented exactly by an undirected graph. The algorithm is evaluated on simulated datasets.

## 1 INTRODUCTION

A graphical probabilistic model makes use of a graph over random variables to encode a dependency model, i.e. a set of marginal and conditional independence relations. Directed acyclic graphs (DAGs) and undirected graphs (UGs) are two popular classes of graphs used to encode dependency models, leading to graphical models known as Bayesian networks and Markov networks (see [9]).

In this paper, we consider the class of graphical models whose structure is a chordal graph, known as the class of decomposable models. A chordal (or triangulated) graph is an undirected graph where every cycle comprising more than three lines has a chord. The class of dependency models defined by chordal graphs is the intersection of the class of DAG dependency models and the class of UG dependency models. The characterization of the independencies of decomposable models has been exploited in [6] in order to construct algorithms for recovering from independence tests the exact chordal structure of a decomposable model, and to build minimal chordal approximations of UG-isomorphic dependency models.

Despite the chordality restriction on the structure, the class of decomposable models is still fairly large and includes, for example, graphical models with undirected tree structure. Also, exact marginalization using the junction-tree algorithm for probabilistic inference over DAGs and UGs is based on the prior transformation of these graphs into a chordal graph [5].

A greedy hill-climbing search algorithm is often used to learn the DAG structure of a Bayesian Network. Different choices of search spaces and neighborhoods connecting the search space are possible. In particular, the search may proceed over the set of Markov equivalence classes of DAG structures by exploiting the inclusion boundary neighborhood (see [1, 4]). Under appropriate assumptions on the scoring criterion and on the data-generating distribution, a greedy algorithm using this inclusion boundary neighborhood returns an inclusion-optimal structure in the limit of a large sample size (see [2] and [3]). Unfortunately, the size of the inclusion boundary of an equivalence class of a DAG structure is in the worst case exponential in the number of variables, which may prevent the application of this strategy in domains with a large number of variables.

The notion of inclusion boundary neighborhood can also be defined over sets of chordal graphs (see Section 2). In this context, its size is bounded from above by the square of the number of variables (pairs of vertices) and it can be computed easily. In [7], this neighborhood is used to learn the chordal structure of a decomposable Gaussian model with a Monte Carlo procedure.

In this paper, we investigate the optimality properties of the greedy hill-climbing search algorithm using the inclusion boundary neighborhood to learn a chordal structure. We describe a local asymptotic consistency property of scoring criteria that ensures that a greedy search will produce an inclusion-optimal chordal structure when the independence relations holding in the data-generating distribution can be represented exactly by an undirected graph. We conjecture that this property still holds when the independencies of the data-generating distribution can be represented exactly by a directed acyclic graph. Hence, we suggest that inclusion boundary based learning of chordal models is an interesting avenue for leveraging learning of graphical models to domains with large numbers of variables.

The rest of the paper is organized as follows. Section 2 defines precisely the mechanism by which an undirected graph encodes a dependency model. It also defines and discusses the notions of inclusion-optimality and inclusion boundary. Section 3 introduces a local consistency property for scoring criteria defined over chordal structures and proves that it holds for common criteria such as the BDe score. Our claim that greedy search with the inclusion boundary neighborhood yields inclusion-optimal solutions is proved there. Section 4 presents some experimental results using simulated datasets.

## 2 BACKGROUND

Consider an undirected graph $G = (X, L)$ whose vertex set $X$ is a set of random variables and whose set of undirected edges (i.e. lines) is denoted by $L$. Given disjoints sets $A, B, C \subseteq X$, we say that $A$ and $B$ are separated by $C$ in $G$ if all paths between a vertex in $A$ and a vertex in $B$ go through at least one vertex in $C$. The dependency model encoded by $G$ consists of the set of marginal and conditional independence relations $A \perp B|C$ such that $A$ and $B$ are separated by $C$ in $G$. In the sequel, we sometimes identify an undirected graph and its dependency model; when we want to distinguish them we will denote by $I(G)$ the dependency model encoded by the graph $G$.

As mentioned in the introduction, a chordal graph is an undirected graph where every cycle more than three lines long has a chord (see Figure 1).

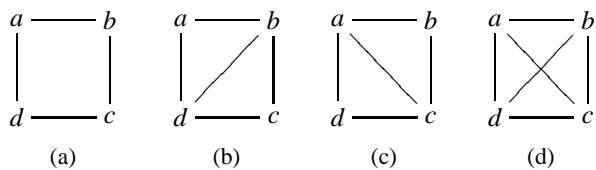

(a)      (b)      (c)      (d)

Figure 1: (a) is undirected, but not chordal since it has the chordless cycle $a, b, c, d, a$ of length four. (b), (c) and (d) are all chordal.

Let us define the notion of inclusion-optimality (aka minimal I-mapness in the terminology of [9]) for chordal graphs. Consider a particular dependency model $M_0$. We say that a *chordal* dependency model $M$ is inclusion-optimal for $M_0$ if $M \subseteq M_0$ and there is no *chordal* dependency model $M'$ such that $M \subsetneq M' \subseteq M_0$. This notion has a simple graphical interpretation: a chordal graph $G$ encodes an inclusion-optimal dependency model for $M_0$ if, and only if, (a) it does not encode any independence assumption that does not hold in $M_0$ and (b) all its proper chordal subgraphs[1] encode such an incorrect independence assumption. For example, suppose that Figure 1(a) encodes $M_0$. Then, the graphs of Figure 1(b) and Figure 1(c) are both inclusion-optimal chordal graphs with respect to $M_0$, while the graph of Figure 1(c) is not. As a special case, note that any chordal graph model is the *unique* chordal dependency model which is inclusion-optimal for itself.

To conclude this section, let us present the notion of inclusion boundary in the context of chordal graphs. The inclusion boundary of a chordal graph $G$ is the set of chordal graphs $H$ satisfying

- $I(G) \subsetneq I(H)$ and there is no chordal graph $K$ such that $I(G) \subsetneq I(K) \subsetneq I(H)$, or

- $I(H) \subsetneq I(G)$ and there is no chordal graph $K$ such that $I(H) \subsetneq I(K) \subsetneq I(G)$.

It is straightforward to describe graphically the inclusion boundary of a chordal graph $G$: it consists of the chordal graphs that differ from $G$ by the addition or removal of a single line. This is a consequence of the fact that, for any two chordal graphs $G, H$ such that $H$ is a subgraph of $G$, there exists a sequence of chordal graphs $K_0, \ldots, K_n$ such that $K_0 = H$, $K_n = G$ and $K_{i+1}$ is obtained from $K_i$ by adding a single line (see [7]).

## 3 INCLUSION-OPTIMALITY OF GREEDY SEARCH

In this section, we first introduce a property of *local* consistency for scoring criteria defined over chordal graphs. Then, we show that if a scoring criterion defined over DAG dependency models is decomposable and consistent in the classical sense (see, e.g. [8]), then it is also locally consistent when restricted to chordal dependency models. Finally, we prove the claim that a greedy hill-climbing search using the inclusion boundary neighborhood and a consistent and locally consistent scoring criterion returns an inclusion-optimal chordal graph when the dependency model of the data-generating distribution can be encoded exactly by an undirected graph.

Following the terminology of [3], we say that a scoring criterion score(·) for chordal graphs is *locally* consistent for a dependency model $I$ if, for any pair of vertices $a, b$ and chordal graphs $G, H$ such that $H$ is obtained from $G$ by removing $a - b$, we have

1. $a \perp b | ne_G(a) \cap ne_G(b) \in I \Rightarrow \text{score}(H) > \text{score}(G)$,

2. $a \perp b | ne_G(a) \cap ne_G(b) \notin I \Rightarrow \text{score}(G) > \text{score}(H)$,

---
[1]We say that $G' = (X', L')$ is a (proper) subgraph of $G = (X, L)$, iff $X' = X$ and $L \subsetneq L'$, i.e. $L$ is a (proper) subset of $L'$.

where $ne_K(a)$ denotes the sets of neighboring (i.e. adjacent) vertices of $a$ in $K$.

Recall that a scoring criterion score(·) for a DAG dependency model encoded by $G$ is decomposable if it can be written as a sum of terms that depend each on only one vertex and its parents, i.e.

$$\text{score}(G) = \sum_{v \in V} f(v, pa_G(v)). \quad (1)$$

The following proposition states that a consistent and locally consistent scoring criterion for chordal dependency models can be obtained from a consistent and decomposable scoring criterion for DAG dependency models. Moreover, it allows us to compute the score difference between neighboring chordal graphs incrementally.

**Proposition 1.** *If* score(·) *is a scoring criterion over DAG dependency models that is decomposable and consistent for a dependency model I, then it is locally consistent for I when restricted to chordal graphs and*

$$\text{score}(G) - \text{score}(H) = f(b, \{a\} \cup (ne_G(a) \cap ne_G(b))) \\ - f(b, ne_G(a) \cap ne_G(b)), \quad (2)$$

*for chordal graphs G and H such that H is obtained from G by removing the line $a - b$.*

PROOF. Consider two chordal graphs $G$ and $H$ such that $H$ is obtained from $G$ by removing the line $a - b$. Since $H$ is chordal and does not have $a-b$, the subgraph of $H$ induced[2] by $ne_H(a) \cap ne_H(b) = ne_G(a) \cap ne_G(b)$ is complete. Hence, the subgraph of $G$ induced by $\{a, b\} \cup (ne_G(a) \cap ne_G(b))$ is complete. If $o_1, \ldots, o_k$ is any ordering of $ne_G(a) \cap ne_G(b)$, there exists a perfect ordering $o$ of $G$ starting with $a, o_1, \ldots, o_k, b$. Let $K$ be the DAG obtained from directing the lines of $G$ according to $o$ and let $L$ be the DAG obtained from $K$ by removing $a \to b$. Note that $K$ and $L$ have no v-structure, score($G$) = score($K$), score($H$) = score($L$), $pa_L(b) = ne_G(a) \cap ne_G(b)$ and $pa_K(b) = \{a\} \cup pa_L(b)$. By decomposability of score(·), we thus have

$$\text{score}(G) - \text{score}(H) = f(b, \{a\} \cup (ne_G(a) \cap ne_G(b))) \\ - f(b, ne_G(a) \cap ne_G(b)). \quad (3)$$

Let $A$ be a complete DAG obtained by orienting the lines of a complete undirected graph according to a vertex ordering starting with $a, o_1, \ldots, o_k, b$ and let $B$ be the DAG obtained from $A$ by removing $a \to b$. We have $pa_B(b) = pa_L(b)$, $pa_A(b) = pa_K(b)$, dim($B$) < dim($A$), $I(A) = \emptyset$ and

$$I(B) = \{a \perp b | ne_G(a) \cap ne_G(b), b \perp a | ne_G(a) \cap ne_G(b)\}. \quad (4)$$

By decomposability of score(·), we have

$$\text{score}(A) - \text{score}(B) = \text{score}(K) - \text{score}(L). \quad (5)$$

---
[2]The subgraph of $G = (X, L)$ induced by $X' \subseteq X$ is the graph $G' = (X', L')$, where $L' = L \cap (X' \times X')$.

By consistency of score(·) for $I$, the restriction of score(·) over chordal graphs is thus also locally consistent for $I$. ∎

In practice, scoring criteria over DAG dependency models only satisfy the consistency property asymptotically in the limit of a large sample size. When restricted to chordal dependency models, such scoring criteria will thus only be locally consistent asymptotically.

### 3.1 Optimality for UG target dependency models

The main result of this paper can now be stated. Its proof relies on results presented in the appendix.

**Proposition 2.** *If* score(·) *is a scoring criterion for chordal graphs that is consistent and locally consistent for a graph-isomorph dependency model I, then local optima of* score(·) *with respect to the inclusion boundary neighborhood are inclusion-optimal for I.*

PROOF. Let $G$ be a local optimum of score(·) with respect to the inclusion boundary neighborhood. Let us show by contradiction that $I(G) \subseteq I$. Suppose that $I(G) \setminus I \neq \emptyset$. Since $I$ satisfies the symmetry, decomposition and intersection properties (see Proposition 3 in the appendix), there exist vertices $a$ and $b$ such that $a \perp b | V \setminus \{a, b\} \in I(G) \setminus I$. Hence, $G$ does not have the line $a - b$. Let us discuss separately the cases where the addition of $a - b$ to $G$ results in a graph $H$ which is chordal and the cases where the resulting graph $H$ is not chordal.

Suppose that $H$ is chordal. Then, $H$ is in the inclusion boundary of $G$. By strong union, $a \perp b | V \setminus \{a, b\} \notin I$ implies that $a \perp b | ne_H(a) \cap ne_H(b) \notin I$. By local consistency, we thus have score($H$) > score($G$) and $G$ is not a local optimum.

Suppose that $H$ is not chordal. There exists a chordless cycle in $H$ of length $\geq 4$. Consider the set of chordless cycles in $H$ of maximum length $m \geq 4$ and the corresponding set of paths in $G$ between $a$ and $b$ of length $n = m - 1 \geq 3$. Let $A_0 = \{a\}$, $A_n = \{b\}$ and, for $i = 1, \ldots, n - 1$, let $A_i$ be the set of vertices that can be reached starting from $a$ by hopping along $i$ lines on one of the above paths between $a$ and $b$. By strong union, $a \perp b | V \setminus \{a, b\} \notin I$ implies that $a \perp b | A_{n-1} \notin I$. By Lemma 4 (see the appendix), there thus exists $i \in \{1, \ldots, n - 1\}$ such that $A_{i-1} \perp A_{i+1} | A_i \notin I$ or $a \perp A_i \in I$. Let us discuss the two possibilities separately. First, suppose that $A_{i-1} \perp A_{i+1} | A_i \notin I$. By composition, there exist $u \in A_{i-1}$ and $v \in A_{i+1}$ such that $u \perp v | A_i \notin I$. By chordality of $G$, note that each set $A_i$ induces a complete subgraph. Hence there is a cycle of length $m$ without chord passing through $u$ and $v$ in $H$ and thus no line between $u$ and $v$ in $G$. By maximality of this cycle, adding $u - v$ to $G$ results in a chordal graph $H'$ in the inclusion boundary of $G$. Since $ne_G(u) \cap ne_G(v) \subseteq A_i$, we have $u \perp v | ne_G(u) \cap ne_G(v) \notin I$ by strong union. By local consistency, we thus have score($H'$) > score($G$) and $G$ is not a local optimum. Second, suppose that $a \perp A_i \in I$ for some $i \in \{1, \ldots, n-1\}$ and consider any vertex $u \in A_i$. By decomposition, we have $a \perp u \in I$. There exists a path $p_1, \ldots, p_k$ in $G$ between $p_1 = a$ and $p_k = u$ where no line is a chord. By transitivity, $a \perp u \in I$ implies that $p_j \perp p_{j+1} \in I$ for some $j \in \{1, \ldots, k-1\}$. Since the line $p_j - p_{j+1}$ is not a chord, the graph $H'$ obtained from $G$ by removing $p_j - p_{j+1}$ is chordal and $H'$ is in the inclusion boundary of $G$. By strong union $p_j \perp p_{j+1} \in I$ implies that $p_j \perp p_{j+1} | ne_G(p_j) \cap ne_G(p_{j+1}) \in I$. By local consistency, we thus have $score(H') > score(G)$ and $G$ is not a local optimum.

To conclude the proof, let us show by contradiction that there is no chordal graph $H$ such that $I(G) \subsetneq I(H) \subseteq I$. Since $I(G) \subsetneq I(H)$, there exists $K$ in the inclusion boundary of $G$ such that $I(G) \subsetneq I(K) \subseteq I(H)$. Also, we have $dim(K) < dim(G)$. By consistency, we thus have $score(K) > score(G)$ and $G$ is not a local optimum. ∎

Note that Proposition 2 applies to all local maxima. The greedy search may thus start at any chordal graph and will return an inclusion-optimal chordal graph under the hypotheses of the proposition.

### 3.2 Extension to other graphical dependency models

Although we have not been able to prove it, we suspect that Proposition 2 still holds when the target dependency model $I$ can no longer be represented perfectly by an undirected graph, but rather by a DAG.

However, as illustrated by the graphs given in Figure 2, the proposition no longer holds when $I$ is encoded by a DAG structure with hidden variables.

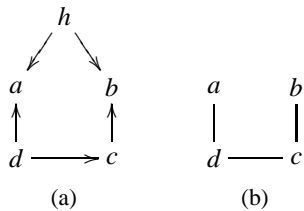

Figure 2: Suppose that $I$ is the set of independence relations over the variables $\{a, b, c, d\}$ encoded by the DAG given in (a). Such a graph does not encode the relation $a \perp b|c$, while the chordal graph given in (b) does encode it, and is thus not inclusion-optimal for $I$. The neighbors of the chordal graph are obtained by adding $a - c$, adding $b - d$, removing $a - d$, removing $b - c$, or removing $c - d$. Using the local consistency property, one can see that each operation decreases the score. The chordal graph is thus a local maximum.

## 4 EXPERIMENTAL RESULTS

This section describes the experiments performed to assess the learning algorithm. The following settings were considered to generate simulated datasets:

- 20 or 50 binary random variables,

- a generating distribution with a chordal structure or a DAG structure.

For each setting, 30 data-generating distributions were selected with random parameters and random structure. DAG structures were drawn randomly with at most 5 parents per variable. Chordal structures were obtained by first drawing DAG structures with at most 3 parents and then chordalizing them with a greedy minimum fill-in algorithm. For each distribution, 30 independent datasets of $10^2$, $10^3$, $10^4$ and, in the case of 20 variables, $10^5$ observations were generated. For each dataset, we learned a chordal structure with the greedy search algorithm using the inclusion boundary. Also, we learned a DAG structure with the greedy search algorithm using the neighborhood obtained by legal arrow additions, removals and reversals. In both case, the BDeu scoring criterion with an equivalent sample size of 1 was used and the search was started at the empty structure. To measure KL divergences with the data-generating distribution, we estimated the parameters of the learned chordal structure, of the learned DAG structure and of the data-generating structure with the Bayesian approach corresponding to our choice of score. Then, for each dataset, the following quantities were measured:

- the dimension, i.e. number of independent parameters, of the data-generating structure,

- the dimension of the learned chordal model,

- the dimension of the learned DAG structure,

- the KL divergence from the distribution with learned parameters and learned chordal structure to the data-generating distribution,

- the KL divergence from the distribution with learned parameters and learned DAG structure to the data-generating distribution,

- the KL divergence from the distribution with learned parameters and data-generating structure to the data-generating distribution,

- in the case of chordal data-generating structure, the number of false positive lines, i.e the lines in the learned chordal structure but not in the data-generating structure, and the number of false negative lines, i.e. the lines in the data-generating structure but not the learned chordal structure.

The KL divergences between a learned distribution $p$ and a target data-generating distribution $g$ were estimated on a dataset $D$ of $10^4$ observations drawn independently of the observations used for learning, according to the following equation:

$$KL(g \parallel p) = |D|^{-1} \sum_{i=1}^{|D|} \ln\left(\frac{P_g(X^i)}{P_p(X^i)}\right), \qquad (6)$$

where $X^i$ denotes the $i$th observation of the test dataset $D$.

Box plots of the results are given in Figure 3 to Figure 12. They are qualitatively similar for data-generating distributions with chordal or DAG structures. Depending on the datasets sizes, one can distinguish three phases. A first phase where the learned chordal model exhibits a lower KL divergence and a lower dimension than the other models. A transition phase where the divergences and dimensions have close values. A final phase where the learned chordal model has a higher dimension and higher divergence, although the divergence tends to decrease. The first phase is expected: the model with correct structure overfits the data, while the learned model benefits from the use of a Bayesian scoring criterion that favors small structures. As the number of observations increases and we enter the third phase, the model with correct structure dominates. However, the model with learned chordal structure seems able to adapt and the difference in divergence keeps decreasing, as an expected consequence of the inclusion-optimality property.

Consider the case of data-generating distributions with chordal structures. As expected again, the number of false positive and false negative lines tends to decrease. Also, note that the number of false positives is in general much lower than the number of false negatives. This is probably due to the fact that the Bayesian score is naturally conservative and gives a high score only to independence relations that are well supported by the data.

## 5 CONCLUSION

In this paper, we discussed the optimality properties of a greedy hill-climbing algorithm using the inclusion boundary neighborhood to learn the structure of a chordal graphical model. We proved that such an algorithm will asymptotically return an inclusion-optimal chordal structure if the scoring criterion is consistent and locally consistent and the dependency model of the data-generating distribution can be represented exactly by an undirected graph. Our experimental results show the practical interest of this algorithm in the context of problems where the number of variables is large and their dependency structure is sufficiently complex, be it UG-faithful or DAG-faithful.

Further theoretical work should address the extension of the above optimality property with respect to more general

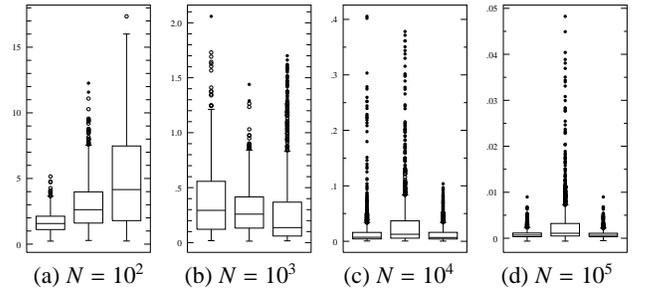

(a) $N = 10^2$    (b) $N = 10^3$    (c) $N = 10^4$    (d) $N = 10^5$

Figure 3: Estimated KL divergences to a data-generating distribution $g$ with chordal structure over 20 binary random variables. For each sample size, the leftmost plot measures the divergence $KL(g \parallel p)$ from the distribution $p$ with learned chordal structure and parameters, the middle plot measures the divergence $KL(g \parallel q)$ from the distribution $q$ with learned DAG structure and parameters, and the rightmost plot measures $KL(g \parallel r)$ from the distribution $r$ with data-generating structure and learned parameters.

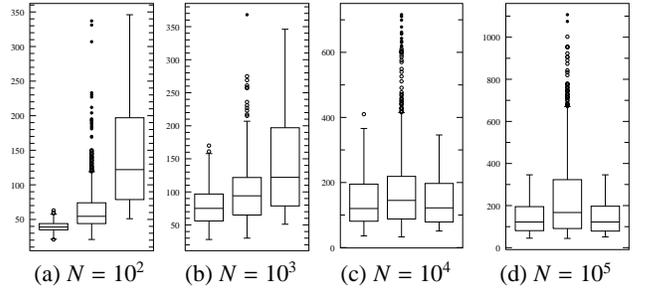

(a) $N = 10^2$    (b) $N = 10^3$    (c) $N = 10^4$    (d) $N = 10^5$

Figure 4: Dimensions with 20 binary random variables and a data-generating distribution with chordal structure. For each sample size, the leftmost plot measures the dimension $d(p)$ of the learned chordal structure, the middle plot measures the dimension $d(q)$ of the learned DAG structure, and the rightmost plot measures the dimension of the data-generating structure $d(g)$.

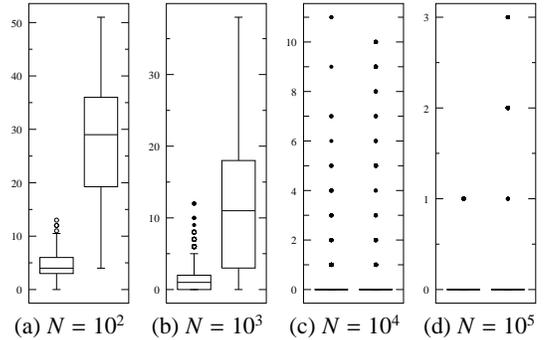

(a) $N = 10^2$    (b) $N = 10^3$    (c) $N = 10^4$    (d) $N = 10^5$

Figure 5: Number of false positive and false negative lines with 20 binary random variables and a data-generating distribution with chordal structure

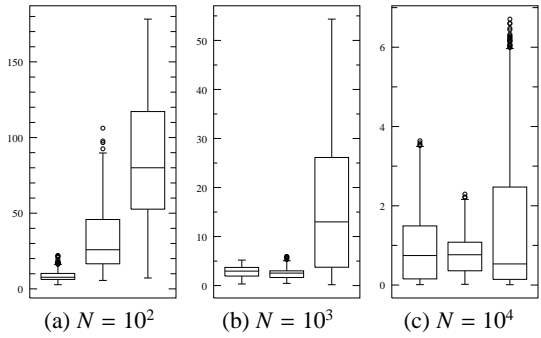

(a) $N = 10^2$    (b) $N = 10^3$    (c) $N = 10^4$

Figure 6: Estimated KL divergences with 50 binary random variables and a data-generating distribution with chordal structure (the layout is the same as in Figure 3).

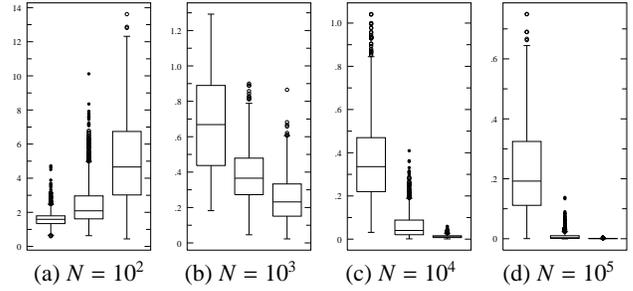

(a) $N = 10^2$    (b) $N = 10^3$    (c) $N = 10^4$    (d) $N = 10^5$

Figure 9: Estimated KL divergences with 20 binary random variables and a data-generating distribution with DAG structure (the layout is the same as in Figure 3).

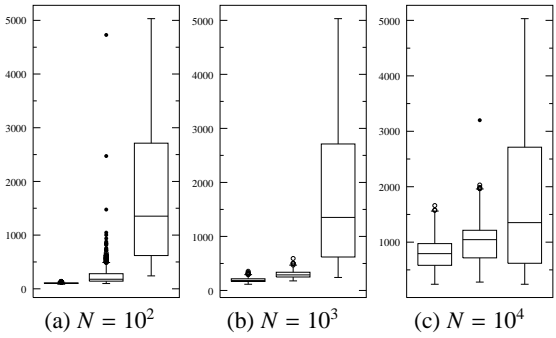

(a) $N = 10^2$    (b) $N = 10^3$    (c) $N = 10^4$

Figure 7: Dimensions with 50 binary random variables and a data-generating distribution with chordal structure (the layout is the same as in Figure 4).

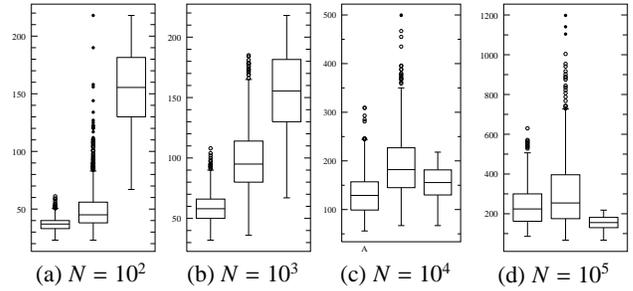

(a) $N = 10^2$    (b) $N = 10^3$    (c) $N = 10^4$    (d) $N = 10^5$

Figure 10: Dimensions with 20 binary random variables and a data-generating distribution with DAG structure (the layout is the same as in Figure 4).

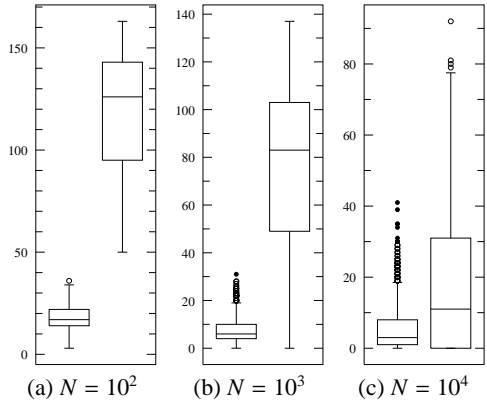

(a) $N = 10^2$    (b) $N = 10^3$    (c) $N = 10^4$

Figure 8: Number of false positive and false negative lines with 50 binary random variables and a data-generating distribution with chordal structure (the layout is the same as in Figure 5).

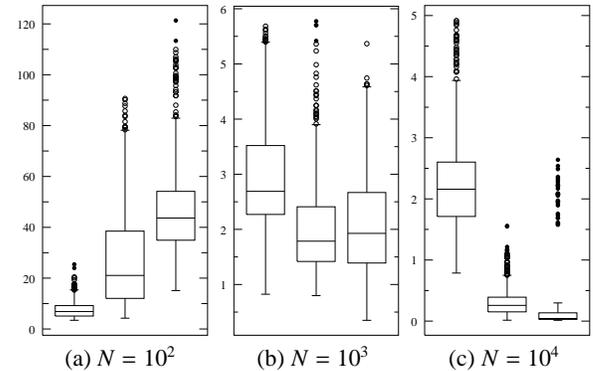

(a) $N = 10^2$    (b) $N = 10^3$    (c) $N = 10^4$

Figure 11: Estimated KL divergences with 50 binary random variables and a data-generating distribution with DAG structure (the layout is the same as in Figure 3).

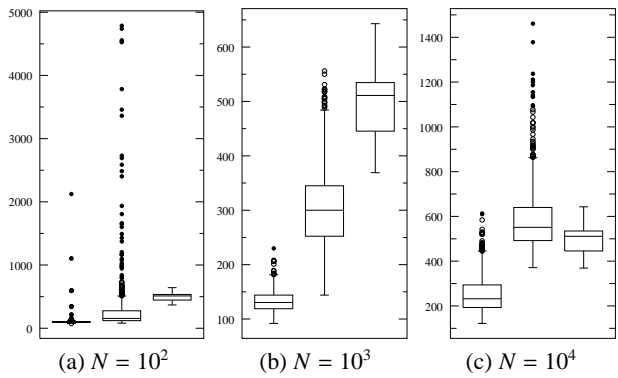

(a) $N = 10^2$  (b) $N = 10^3$  (c) $N = 10^4$

Figure 12: Dimensions with 50 binary random variables and a data-generating distribution with DAG structure (the layout is the same as in Figure 4).

conditions on the data-generating distributions. In particular, we believe that our optimality properties can be extended to the case where the latter is DAG faithful.

From a more practical side, it would be interesting to carry out a more in-depth empirical investigation of inclusion boundary search of chordal models with respect to inclusion boundary search in the larger space of Markov equivalence classes of DAG structures. Also, suitable combinations of these two approaches might lead to further progress for learning graphical models over large numbers of variables.

We believe also that it is of interest to further investigate the possible uses of the greedy search of inclusion optimal chordal models in the context of designing efficient approximate inference algorithms, in particular in the framework of variational approximations [10].

**Acknowledgments**

This work presents research results of the Belgian Network BIOMAGNET (Bioinformatics and Modeling: from Genomes to Networks), funded by the Interuniversity Attraction Poles Programme, initiated by the Belgian State, Science Policy Office. Vincent Auvray is supported by the "Action de recherche concertée" BIOMOD funded by the French Speaking Community of Belgium. We also want to thank Dr. R. Castelo for very fruitful discussions in September 2007, which encouraged us to direct our work on inclusion-boundary driven learning by considering the space of chordal graphs. We thank the anonymous reviewers for their very useful comments about our work.

## APPENDIX

**Proposition 3 (from [9]).** *A dependency model I can be represented exactly by an undirected graph if, and only if, it satisfies the following properties:*

1. *symmetry*
$$X \perp Y | Z \in I \Leftrightarrow Y \perp X | Z \in I,$$

2. *decomposition*
$$X \perp Y \cup W | Z \in I \Rightarrow X \perp Y | Z \in I \wedge X \perp W | Z \in I,$$

3. *intersection*
$$X \perp Y | Z \cup W \in I \wedge X \perp W | Z \cup Y \in I$$
$$\Rightarrow X \perp Y \cup W | Z \in I,$$

4. strong union

$$X \perp Y | Z \in I \Rightarrow X \perp Y | Z \cup W \in I,$$

5. transitivity

$$X \perp Y | Z \in I \Rightarrow X \perp \gamma | Z \in I \vee \gamma \perp Y | Z \in I,$$

where W, X, Y, and Z are disjoints subsets of vertices and $\gamma$ is a singleton vertex.

**Lemma 4.** *Let I be a graph-isomorph dependency model. If $A_0, \ldots, A_n$ ($n \geq 3$) are sets of vertices such that $A_0 = \{x\}$, $A_n = \{y\}$, and $A_{i-1} \perp A_{i+1} | A_i \in I$ for $i = 1, \ldots, n-1$, then we have*

$$x \perp A_1 \in I \vee \cdots \vee x \perp A_{n-1} \in I \vee x \perp y | A_{n-1} \in I.$$

PROOF. By transitivity and symmetry, we have

$$A_1 \perp A_3 | A_2 \in I \Rightarrow x \perp A_1 | A_2 \in I \vee x \perp A_3 | A_2 \in I.$$

By intersection, we have

$$x \perp A_1 | A_2 \in I \wedge x \perp A_2 | A_1 \in I \Rightarrow x \perp A_1 \cup A_2 \in I.$$

Hence, we have

$$x \perp A_1 \cup A_2 \in I \vee x \perp A_3 | A_2 \in I.$$

Repeating these steps, we obtain

$$x \perp A_1 \cup A_2 \in I \vee x \perp A_2 \cup A_3 \in I \vee \ldots$$
$$\vee x \perp A_{n-2} \cup A_{n-1} \in I \vee x \perp y | A_{n-1} \in I.$$

By decomposition, we thus have

$$x \perp A_1 \in I \vee \cdots \vee x \perp A_{n-1} \in I \vee x \perp y | A_{n-1} \in I. \quad \blacksquare$$